\documentclass[letterpaper, 10 pt, conference]{ieeeconf}  
\usepackage{caption}
\usepackage{amssymb}
\usepackage{amsmath}
\usepackage{graphicx}
\usepackage{xspace}
\usepackage{xcolor}
\usepackage[acronym]{glossaries}
\usepackage{booktabs}
\usepackage{tabularx}
\newcommand{\finetact}{{TranTac}\xspace}

\newacronym{cnn}{CNN}{convolutional neural network}
\newacronym{imu}{IMU}{inertial measurement unit}
\newacronym{ast}{AST}{audio spectrogram transformer}

\IEEEoverridecommandlockouts                              

\title{TranTac: Leveraging Transient Tactile Signals for Contact-Rich Robotic Manipulation}

\author{Yinghao Wu$^{1}$, Shuhong Hou$^{1}$, Haowen Zheng$^{1}$, Yichen Li$^{1}$, Weiyi Lu$^{1}$, Xun Zhou$^{1}$, Yitian Shao$^{1,2}
$\thanks{$^{1}$School of Computer Science and Technology, Harbin Institute of Technology, Shenzhen, Shenzhen, China (correspondence: shaoyitian@hit.edu.cn).}
\thanks{$^{2}$State Key Laboratory of Smart Farm Technologies and Systems, Harbin, China.}}

\begin{document}

\maketitle
\thispagestyle{empty}
\pagestyle{empty}

\begin{abstract}
Robotic manipulation tasks such as inserting a key into a lock or plugging a USB device into a port can fail when visual perception is insufficient to detect misalignment. In these situations, touch sensing is crucial for the robot to monitor the task's states and make precise, timely adjustments.
Current touch sensing solutions are either insensitive to detect subtle changes or demand excessive sensor data.
Here, we introduce \finetact, a data-efficient and low-cost tactile sensing and control framework that integrates a single contact-sensitive 6-axis inertial measurement unit within the elastomeric tips of a robotic gripper for completing fine insertion tasks. 
Our customized sensing system can detect dynamic translational and torsional deformations at the micrometer scale, enabling the tracking of visually imperceptible pose changes of the grasped object. By leveraging transformer-based encoders and diffusion policy, \finetact can imitate human insertion behaviors using transient tactile cues detected at the gripper's tip during insertion processes. These cues enable the robot to dynamically control and correct the 6-DoF pose of the grasped object. 
When combined with vision, \finetact achieves an average success rate of 79\% on object grasping and insertion tasks, outperforming both vision-only policy and the one augmented with end-effector 6D force/torque sensing.
Additionally, TranTac's contact localization performance is validated through tactile-only insertion tasks, where the inserted object and slot are initially misaligned by 1 to 3 mm, achieving an average success rate of 88\%. We assess the generalizability by training \finetact on a single prism-slot pair and testing it on unseen data, including a USB plug and a metal key, and find that the insertion tasks can still be completed with an average success rate of nearly 70\%.
The proposed framework may inspire new robotic tactile sensing systems for delicate manipulation tasks.
\end{abstract}

\section{INTRODUCTION}

\begin{figure}[ht]
    \centering
    \includegraphics[width=0.5\textwidth]{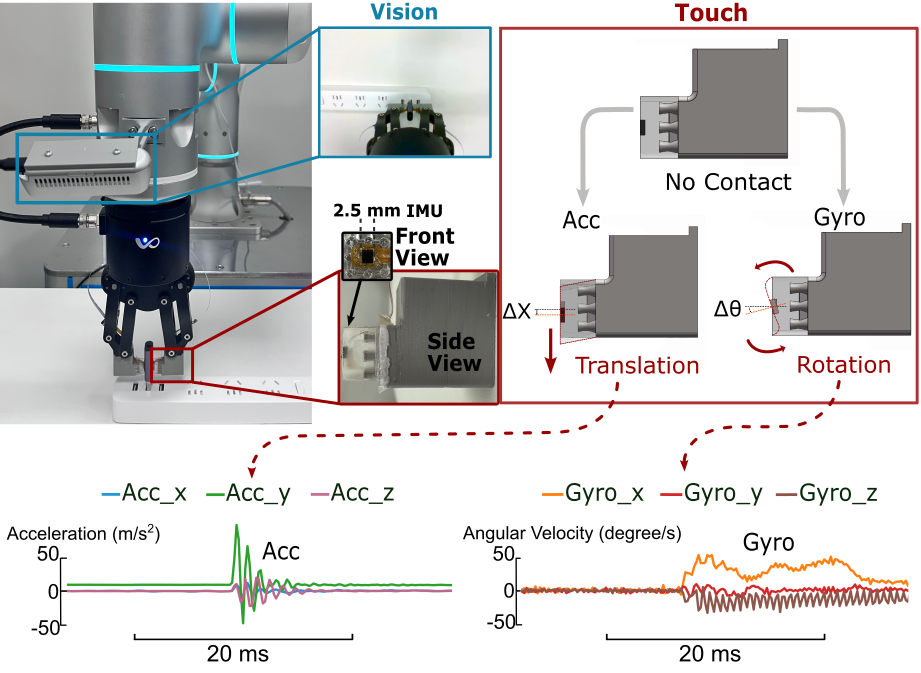}
    \caption{Overview of \finetact, a tactile sensing and control framework for fine robotic manipulation. The framework combines visual input from a wrist-mounted camera with tactile sensing from fingertip-embedded \gls{imu}.
    When the grasped objects interacts with the environment, the embedded \gls{imu}s in the gripper's tip capture the subtle translation and torsion of the elastomeric tip, which produces 3-axis acceleration (ACC) signals and 3-axis angular velocity (Gyro) signals respectively. Both types of signals are captured by the \gls{imu}s and provide fine-grained feedback for precise 6-DoF control of the robot end-effector. 
}
    \vspace{-1em}
    \label{fig:sensor_overview}
\end{figure}
Recent advancements in vision-language-action models have sparked rapid development of robotic technologies, allowing robots to learn various manipulation skills, such as rearranging tools and delivering objects \cite{brohan2022rt, brohan2023rt}.
However, delicate manipulation remains challenging because vision alone often fails to detect transient, subtle changes in objects.
In addition, visual sensing can be obstructed when handling small objects or operating in complex environments. In these situations, touch sensing is highly desirable.

Many tactile sensing strategies have been proposed \cite{Li2020}, with visuo-tactile sensing emerging as one of the most popular choices due to its high spatial resolution and exceptional reliability \cite{yuan2017gelsight, kuppuswamy2020soft,shimonomura2019tactile}.
However, because these sensors prioritize the capture of detailed spatial information, they suffer from limited responsiveness due to their relatively low sampling rate. This prevents robots from reacting quickly in dynamically changing, contact-rich environments.
To address this, some designs have explored the use of contact microphones~\cite{du2022play,li2023see, thankaraj2023sounds, mejia2024hearing, liu2024maniwav} or event cameras~\cite{rigi2018novel, kumagai2019event, ward2020neurotac, funk2024evetac} to detect transient touch events. However, event cameras are costly, while contact microphones lack directional information.

\begin{table*}[t]
\caption{\textnormal{Comparison of \finetact (ours) with existing tactile sensing methods previously applied to object insertion tasks. Performance indicators are marked to show whether $\downarrow$ low values are better or $\uparrow$ high values are better, with the best value highlighted in bold.}}
\label{tab:tactile_comparison}
\centering
\scriptsize 
\setlength{\tabcolsep}{2.5pt} 
\renewcommand{\arraystretch}{1.1} 
\begin{tabularx}{\textwidth}{lXXXXXX}
\toprule 
Sensor & \finetact (Ours) & GelSlim 3.0~\cite{taylor2022gelslim} & 9DTact~\cite{lin20239dtact} & DIGIT~\cite{lambeta2020digit} & ReSkin/AnySkin~\cite{bhirangi2025anyskin} & 3D-ViTac~\cite{huang20243d} \\
\midrule
$\downarrow$ Size [mm] & \textbf{11$\times$11$\times$8} & 37$\times$80$\times$20 & 33$\allowbreak\times\allowbreak$26$\allowbreak\times\allowbreak$26 & 20$\times$27$\times$18 & 20$\times$20$\times$2 & 48$\times$48$\times$3 \\
$\downarrow$ Cost [\$] &  \textbf{5} & 25 & 6 & 15 & 30 & 20 \\
$\downarrow$ Channels/Pixels &  \textbf{6} & 640$\times$480 & 640$\times$480 & 640$\times$480 & 15 & 16$\times$16$\times$4 \\
$\downarrow$ Data Efficiency (Stream Volume) [KB/s] & 42 & 27648 & 27648 & 55296 & \textbf{12} & 33 \\
$\uparrow$ Sensing Bandwidth [Hz] &  \textbf{3500} & 30 & 30 & 60 & 400 & 32.2 \\
\bottomrule
\end{tabularx}
\vspace{-1em}
\end{table*}

In this paper, we draw inspiration from the dexterity of the human hand, which utilizes dynamic tactile sensing to efficiently track the pose of grasped objects \cite{johansson2009coding}. In particular, a light touch contact on the human fingertip can be readily captured by the tactile sensors embedded in the skin, efficiently encoding contact information through sparse temporal neural coding \cite{shao2020compression}.
Such encoding mechanism is characterized by high temporal fidelity, informing the importance of a wide bandwidth tactile sensing system. Here, our approach focuses on the design of gripper tips, where contact interactions most frequently occur during manipulation.
Given the need for timely tracking and control of the 6-DoF pose of a grasped object, we consider that a sensor capable of detecting both translational and torsional deformations of the gripper is sufficient.
We designed \finetact, a novel robotic sensing and control framework that leverages contact-sensitive, wide-bandwidth, and low-cost 6-axis \gls{imu} sensors to capture subtle and transient deformations of the gripper tip induced by pose variations of the grasped object.
Compared to existing methods, our hardware design is both data-efficient and cost-effective, while retaining the capability to capture dynamic tactile information with high temporal accuracy (Table~\ref{tab:tactile_comparison}).
It learns a visuotactile policy through diffusion of action, enabling fine manipulation (Fig.~\ref{network architecture}).
Temporal features are extracted from instantaneous \gls{imu} data, encoded via transformers, fused with vision features, and fed into a diffusion model to generate 6-DoF poses for subsequent movement steps.
We validate the effectiveness of \finetact through physical experiments, in which the robot grasps and inserts objects of various shapes and materials. 

The key contributions of this paper are as follows:
1) A novel tactile sensing mechanism for robots that leverages temporal tactile information to efficiently detect transient, subtle pose changes in grasped objects.
2) A compact and readily reproducible design of the tactile sensing hardware that utilizes low-cost, off-the-shelf \gls{imu} chip and silicone molding.
3) A robot imitation learning framework based on action diffusion, which integrates \gls{imu}-based tactile data with spatial visual information to perform insertion tasks using a diverse set of objects—from plastic prisms to everyday items like USB plugs and metal keys.


\section{Related Work}\label{sec:relatedworks}
\textbf{Tactile sensing for robot manipulation:}
Diverse tactile sensing designs have been proposed for robot grippers. Force/torque sensing are widely used in industrial robots \cite{cao2021six}, but their bulky size makes them unsuitable for installation at the tip of a robot gripper, limiting their effectiveness in localized tactile sensing.
Tactile sensors that utilize piezoresistive, piezoelectric, or capacitive components can be embedded in robot skin with slim profiles \cite{luo2021learning, egli2024sensorized, huang20243d, kasolowsky2024fine}. 
However, these sensors often require complex fabrication processes and acquisition electronics, and they are susceptible to environmental noise and crosstalk. 
Visuo-tactile sensors \cite{yuan2017gelsight, shimonomura2019tactile} have gained increasing attention due to their excellent spatial resolution and reliability. The deformation of a soft robot gripper tip can be captured by an embedded camera, capturing high-precision data on the contact location and surface texture of the grasped object. However, the camera and required optical path space enlarge the end effector equipped with such sensors. Moreover, the temporal resolution of visuo-tactile sensing is constrained by the camera frame rate, making it difficult to capture high-frequency transient tactile signals at the kilohertz level. 
Acoustic sensors such as piezoelectric contact microphones have been integrated into robotic grippers or affixed to experimental objects to compensate for the response delay of vision-based tactile sensors in detecting contact events, modes and object states \cite{du2022play, li2023see, thankaraj2023sounds, mejia2024hearing, liu2024maniwav}. However, bulky contact microphones are hard to be integrated into gripper tips, and their signals lack directional information and are highly susceptible to ambient acoustic noise \cite{liu2024maniwav}.

\textbf{Peg-in-hole insertion via touch:}
Peg-in-hole problems are classic but challenging benchmarks for robotic manipulation. Early works relied primarily on visual sensing for object localization and alignment, but vision alone often struggles with occlusion and limited accuracy in fine insertion tasks \cite{dahiya2010tactile, kappassov2015tactile}. Recent research has increasingly focused on the integration of visuo-tactile sensors to provide high-resolution local contact information for precise insertion tasks. 
A constraint-based estimation framework has been developed to localize extrinsic contacts using distributed tactile sensing \cite{ma2021extrinsic}. 
Streaming tactile imprints can be leveraged to estimate and correct object pose errors, enabling robust insertion of unknown objects without prior geometric knowledge \cite{dong2021tactile}. Tactile sensing can also support closed-loop control in complex tool-using manipulation tasks by providing accurate real-time pose estimation \cite{shirai2023tactile}.
Visuo-tactile sensing typically processes sequences of tactile images, where information about the pose or extrinsic contact location of the grasped object is implicitly encoded in their variations. This requires extensive sensor data across both spatial and temporal domains, resulting in delayed pose adjustments. In addition, to better capture transient tactile cues, active exploration has been incorporated to estimate contact locations \cite{kim2022active}.

\textbf{Imitation learning for robot manipulation:}
Imitation learning enables robots to acquire complex manipulation skills through demonstrations. Recent approaches employ deep-generative models to preserve the variability of imitated trajectories, allowing for generalization to new scenarios \cite{kim2024openvla, team2024octo, intelligence2025pi}. End-to-end imitation learning has been shown to enable low-cost hardware to perform fine-grained bimanual manipulation tasks based on real-world demonstrations \cite{zhao2023learning}. Diffusion-based generative models have demonstrated potential in visuomotor policy learning, enabling robots to generate diverse and adaptive motion plans \cite{chi2023diffusion}.






	


\section{Bio-inspired Dynamic Sensing for Subtle Touch}
Humans can perform blind insertion tasks by perceiving subtle contact between a grasped object and its surroundings. 
Tactile sensors in the hand play a pivotal role, as they are extremely sensitive to transient skin deformations \cite{johansson2009coding} and can efficiently capture tactile information through sparse but highly precise temporal encoding \cite{tummala2023spatiotemporal, shao2020compression}.
Inspired by this, we designed our \finetact sensing system using a single 6-axis \gls{imu} with a temporal resolution as high as 0.28\,ms. 

This design aims to responsively and efficiently capture the 3-axis translation and 3-axis torsional deformation that occur in the contact region of the gripper tip. 
The sensing efficiency of our design can be compared to emerging visuo-tactile sensing techniques, where the status of grasped object is encoded by high-dimensional spatio-temporal tactile signals recorded by an integrated camera. For example, a 30\,fps, 480p camera produces at least 9000 kilobytes of data per second ($30 \times 640 \times 480$), not to mention that many existing studies utilize cameras with even higher spatial and temporal resolutions. In contrast, our design features a data volume of just 42 kilobytes per second. The detailed comparison between different sensors is provided in Table~\ref{tab:tactile_comparison}.





\subsection{Gripper Tip with 6D Dynamic Tactile Sensing}
\finetact employs a soft elastomeric fingertip with an embedded 6-axis \gls{imu} (ST LSM6DSR iNEMO) to capture subtle translational and torsional deformations of the elastomeric tips (Fig.~\ref{fig:sensor_overview}). The size of the \gls{imu} chip is only $2.5\times 3$\,mm, compact enough to be integrated into gripper tips.
The sensor measures accelerations up to ±16 g and angular velocities up to ±4000 dps. IMU data are streamed via a flat flex cable at 3500 Hz to a compact computing unit (Raspberry Pi 5), which transmits the data in real-time to a circular buffer on the PC for storage and processing by the action generation module.

\begin{figure}[bp]
    \centering
    \includegraphics[width=90mm]{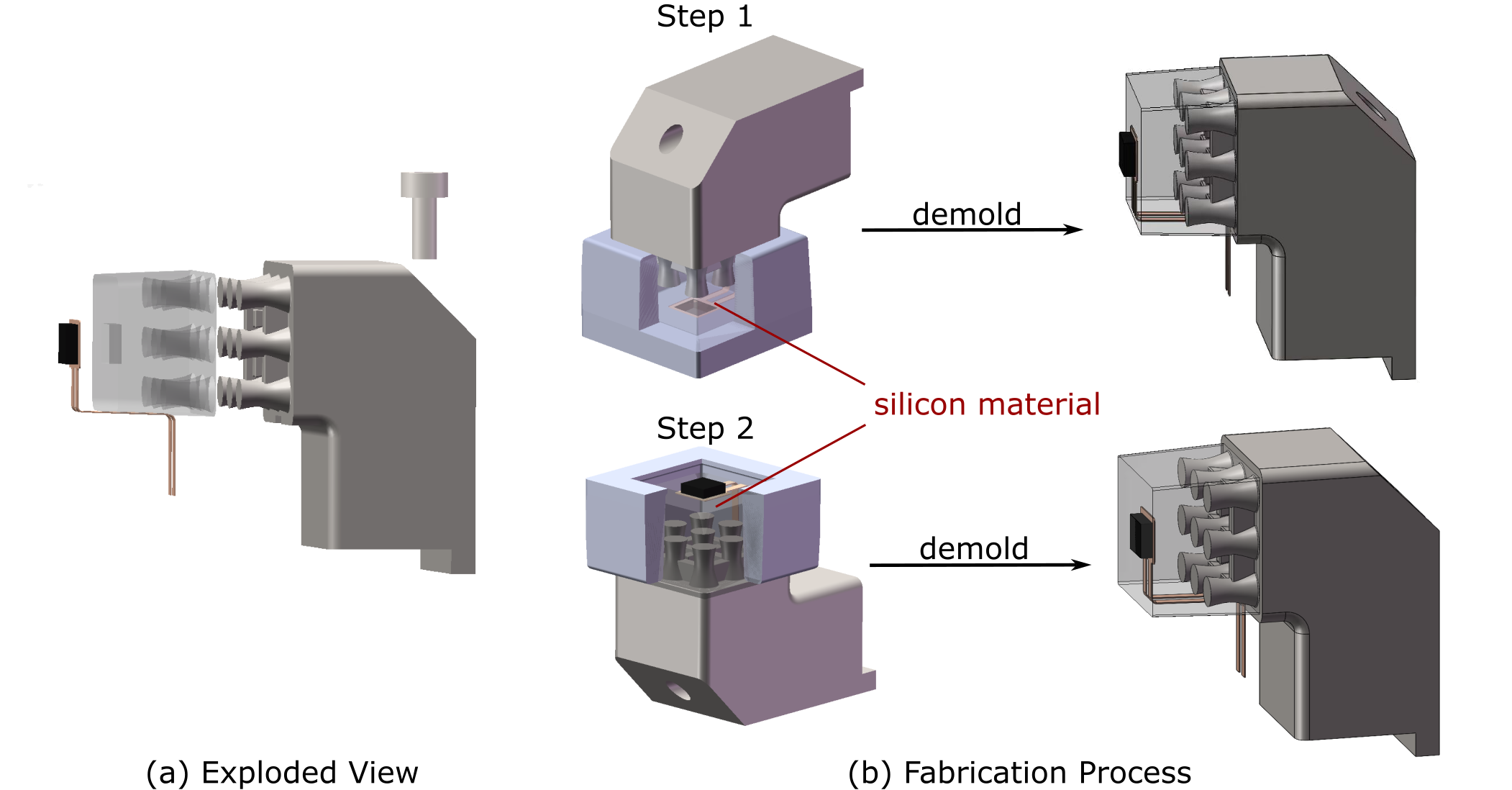}
    \caption{Design of \finetact tactile sensor: a) exploded view and b) fabrication process.}
    \label{fig:fabrition}
    \vspace{-0.5em}
\end{figure}
The elastomeric tip is made from polydimethylsiloxane (PDMS, Dow SYLGARD 184) and is molded onto the protrusion of the 3D-printed fingertip support. 
Fig.~\ref{fig:fabrition} shows the fabrication process. The IMU board is first placed upside down into the slot at the bottom of the mold. The PDMS is then poured to fill the mold, after which the elastomeric tip base is inserted into the silicone-filled slot. The elastomer is then cured at 60\,\textdegree{}C for 120 minutes. Once cured, the assembly is cooled to room temperature and demolded. Then, another mold without a bottom slot is fitted over the solidified silicone fingertip. An additional layer of silicone material is added to cover the IMU. The elastomer is then cured at 60\,\textdegree{}C for another 120 minutes. Finally, after being cooled to room temperature, the mold is removed, completing the fabrication process.

\subsection{Tactile Sensing for Contact Localization}
\begin{figure}[h]
    \centering
    \includegraphics[width=76mm]{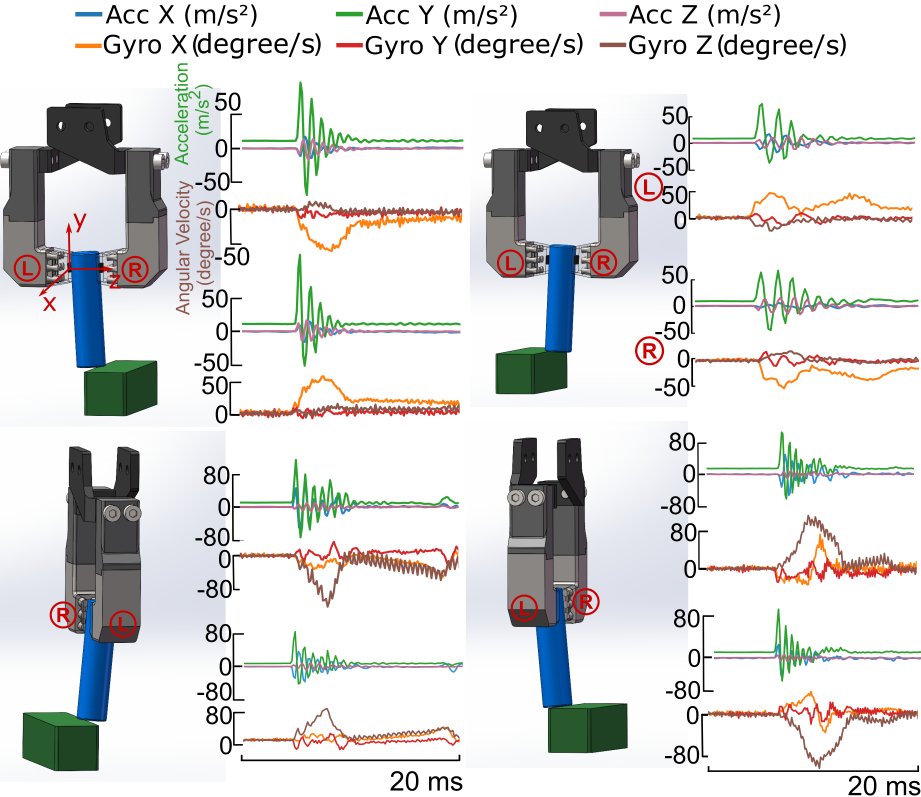}
    \caption{Examples of measured \finetact signals for four contact directions of a grasped object.  
    The top two cases illustrate contact with an edge oriented along the x-axis of the gripper coordinate system, resulting in magnitude differences between the left and right acceleration (Acc) signals and corresponding variations in the gyroscope (Gyro) signals.
    The bottom two cases show contact with an edge along the z-axis of the gripper coordinate system, varying the z-axis Gyro signals, with opposite phase observed between the left and right sensors.}
    \label{fig:direction}
\end{figure}
Fig.~\ref{fig:direction} shows four scenarios where an object grasped by the parallel gripper moves vertically downward to contact an edge in the environment and subsequently lifts off. In each scenario, we visualize the 20\,ms collision signals, which capture the immediate response at contact. 

As demonstrated in the figure, different collision directions produce distinct signal patterns. In the first row, the gripper contacts an edge that is parallel to the sensor plane, resulting in a difference in the magnitude of the left and right acceleration signals along the y-axis and relatively large changes in the gyroscope (Gyro) x-axis signals. In the second row, the gripper contacts an edge perpendicular to the sensor plane, causing the object to rotate within the sensor plane and leading to a significant change in the gyroscope signals along the z-axis. 

\begin{figure}[htbp]
\centering
\includegraphics[width=76mm]{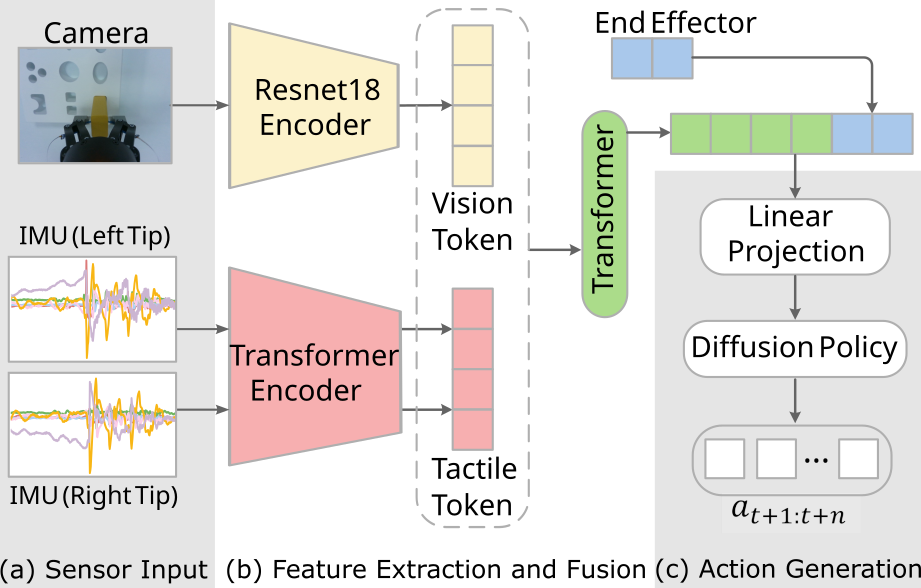}
    \caption{Network architecture of the proposed model. It consists of three parts: a) Sensor input, which contains camera images and two IMU tip signals. b) Feature extraction and fusion, including the concatenation of vision and tactile tokens and a Transformer for multi-modal and temporal feature integration together with the current end-effector state. c) A conditional diffusion strategy to generate robot end-effector poses.}
    \label{network architecture}
    \vspace{-0.5em}
\end{figure}

\subsection{Policy Learning}
Human hands can adjust the pose during tool use by perceiving dynamic translational and torsional deformations of the skin at fingertips~\cite{johansson2009coding}. To enable robots to imitate this human capability, we propose an end-to-end visuotactile policy that maps visual and tactile observations to actions, as illustrated in Fig.~\ref{network architecture}. Our method consists of three critical parts. Fig. 4(a) shows the sensor inputs, including one wrist-mounted camera and two 6-axis IMU sensors at the fingertips. The camera sampling frequency is 24\,Hz, and the IMU sampling rate is 3500\,Hz.
Fig.~\ref{network architecture}(b) illustrates the feature extraction and fusion process from the image and imu signals, and Fig.~\ref{network architecture}(c) depicts the policy learning process for action generation, which leverages the diffusion policy~\cite{chi2023diffusion} conditioned on our tactile representation to generate 6-DoF actions for the robot end-effector.

\subsubsection{Visuotactile Encoding}
We use camera images as visual input and adopt ResNet-18~\cite{he2016deep} as the visual encoder. 
Within each visual frame interval, the 146-frame sequence of 6-axis IMU signals is used as input to the tactile module. Specifically, the IMU signals are projected to 64 dimensions through a single MLP layer, then processed by a transformer to extract features. We use the same IMU encoder to extract left tip tokens and right tip tokens, which are then concatenated and fused with visual tokens using a transformer.
Finally, the fused tokens are concatenated and projected to 512 dimensions, then concatenated with the robot proprioception features to serve as input to the action head. Note that we use the last two time steps' sensor signals as observations.

\subsubsection{Action Generation}
Our action head is a diffusion policy $D_{policy}$~\cite{chi2023diffusion}.
As illustrated in Fig.~\ref{network architecture}, the policy network learns to denoise actions by predicting $\varepsilon_\theta$ from multimodal visual, tactile, and proprioceptive observations $O$. We employ the conventional DDPM~\cite{ho2020denoising} loss function:
\begin{equation}
L = \mathbb{E}_{(O,A_0) \in D_{policy}, k, \varepsilon_k} \left\| \varepsilon_k - \varepsilon_\theta(O, A_0 + \varepsilon_k, k) \right\|^2,
\end{equation}
where $k$ represents the noise schedule step, $\varepsilon_k$ is the ground-truth noise, and $A_0$ represents the ground-truth 16-step future robot trajectories. We predict the noise $\varepsilon_\theta$ through a CNN-based diffusion model with FiLM conditioning~\cite{perez2018film}.



\section{Experiments}
In this section, we investigate extensive contact-rich insertion experiments to demonstrate the effectiveness and generalization of using tactile cues from the insertion process to help policy learning. These experiments are designed to answer the following questions:

\begin{itemize}
    \item Does \finetact improve policy performance in contact-rich tasks compared to both pure vision-based policies and robot end-effector 6D force sensor? (Experiment 1)
    \item Can \finetact correctly identify the contact location and the direction of adjustment? (Experiment 2)
    \item Can tactile cues captured by \finetact generalize to different objects? (Experiment 2)
\end{itemize}

\subsection{Environment setup}
\begin{figure*}[htbp]
    \centering
    \includegraphics[width=162mm]{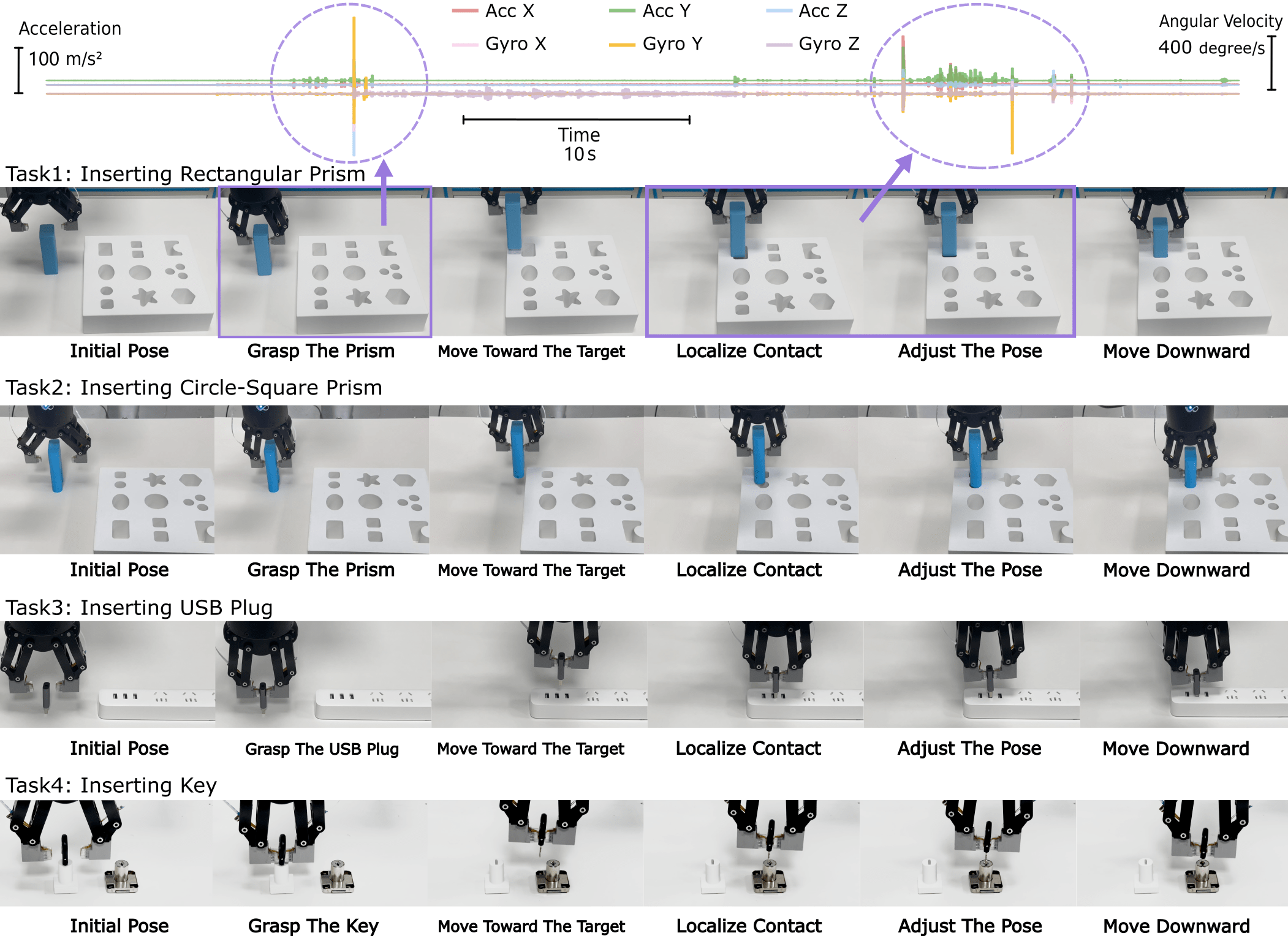}
    \caption{Experimental objects used in the insertion experiments, including 3D printed plastic prisms and insertion board, USB connector-hub pair, and key-lock pair. The top panel displays the transient tactile signals captured by \finetact, highlighting key events during the grasping and insertion process.}
    \label{fig:experimental_objects}
    \vspace{-0.5em}
\end{figure*}
The hardware used in the experiment is shown in Fig.~\ref{fig:sensor_overview}, including a 7-DoF robot arm (Flexiv Rizon 4s) integrated with a 6D force/torque sensor and equipped with a gripper (Flexiv Grav), two \finetact sensors mounted on the gripper tips, 
and a RGB-D camera (RealSense D435i) mounted on the robot arm for wrist view. For each task, demonstrations are collected at 24\,Hz via a teleoperation system employing a see-through VR headset~\cite{xue2025reactive}. Learned policies are deployed at a 12\,Hz frequency. We use action chunking with exponential temporal averaging~\cite{zhao2023learning} to produce smoother behavior.

\subsection{Experiment 1: Policy Comparison}
\subsubsection{Task Description}
Fig.~\ref{fig:experimental_objects} shows the experimental objects used in the insertion experiments. We use 3D printable objects from~\cite{luo2025fmb}, including a rectangle and a circle-square prism with insertion slot clearances of only 0.5\,mm. Additionally, we test with two everyday objects, a USB connector-hub pair and a key-lock pair.

\textbf{Plastic prism insertion tasks:} This task requires the robot to insert a 3D-printed object—either a simple rectangular shape or a composite form consisting of a circle and a square—into its designated slot. The arm starts at a random position near the target slot and places the object directly below the gripper. 
The training dataset consists of 40 demonstration trajectories.

\textbf{USB insertion task:} This task requires the robot to plug a USB connector into a specific port on a power hub. The arm starts at a random position near the power hub 
and places the object directly below the gripper. 
The training dataset consists of 40 demonstrations.

\textbf{Key insertion task:} This task requires the robot to insert a metal key into its corresponding lock. The robotic arm begins from a random position to the left of the lock, with the key positioned directly beneath the gripper. The training dataset comprises 40 demonstrations.

Note that during testing, we randomly initialize the gripper location using the same distribution as in training to prevent the network from memorizing the initial robot pose, while maintaining the same relative position between the object and the slot. 

Based on the four tasks above, we evaluated and compared the following three policies. Each policy was tested using 20 rollouts per task to ensure reliable assessment.
\begin{itemize}
    \item \textbf{Vision Only:} ResNet-18 encoder extracts 512-dimensional features from RGB inputs, concatenated with 6D gripper pose (proprioceptive information) as action-conditional features.
    
    \item \textbf{Vision with Force:} 6D force/torque data is projected to 512 dimensions via MLP and fused with ResNet-18 visual features using Transformer attention, with 6D gripper pose concatenated as action-conditional features.
    
    \item \textbf{Vision with \finetact:} Multi-modal architecture (Fig.~\ref{network architecture}) incorporating vision and tactile (\finetact) modalities, with 6D gripper pose as action-conditional features. No force/torque data is used.
\end{itemize}

\subsubsection{Qualitative Analysis}
\begin{table}[bp]
\centering
\caption{\textnormal{Success rates (out of 20) of four insertion tasks for comparing three policies.}}
\label{tab:success_rates}
\scriptsize
\setlength{\tabcolsep}{4pt}
\renewcommand{\arraystretch}{1.1}
\begin{tabular}{lcccc}
\toprule
Policy & Rectangle & Circle-square & USB & Key\\
       & Insertion & Insertion     & Insertion & Insertion\\
\midrule
Vision Only      & 75\% & 60\% & 30\% & 80\%\\
Vision w. Force    & 50\% & 70\% & 40\% & 40\%\\
\textbf{Vision w. \finetact} & \textbf{80\%} & \textbf{80\%} & \textbf{65\%} & \textbf{90\%}\\
\bottomrule
\end{tabular}
\end{table}
Table~\ref{tab:success_rates} shows the experimental success rates. Our results show that \finetact policies outperform both vision-only approaches and vision-based methods augmented with 6D force/torque sensing.
Additionally, we observe that \finetact policies can fit the demonstration datasets more precisely, particularly for contact strategies at hole edges, enabling accurate contact localization and adjustment for more precise insertion operations. In contrast, force-based policies suffer from noise in the robot's end-effector 6D force/torque measurements, which adversely affects performance in fine manipulation tasks. The vision-only policy struggles with occlusion issues, especially in circle-square insertion tasks where the inserted object blocks the view of the square hole, frequently causing the robot to get stuck near the hole entrance and resulting in lower success rates for pure vision approaches.


\subsection{Experiment 2: Generalizability of Tactile-Only Policy}
To validate \finetact's contact localization capability and the generalizability of tactile features, we designed tactile-only policy experiments. 
\subsubsection{Task Description}
\begin{figure}[tp]
    \centering
    \includegraphics[width=78mm]{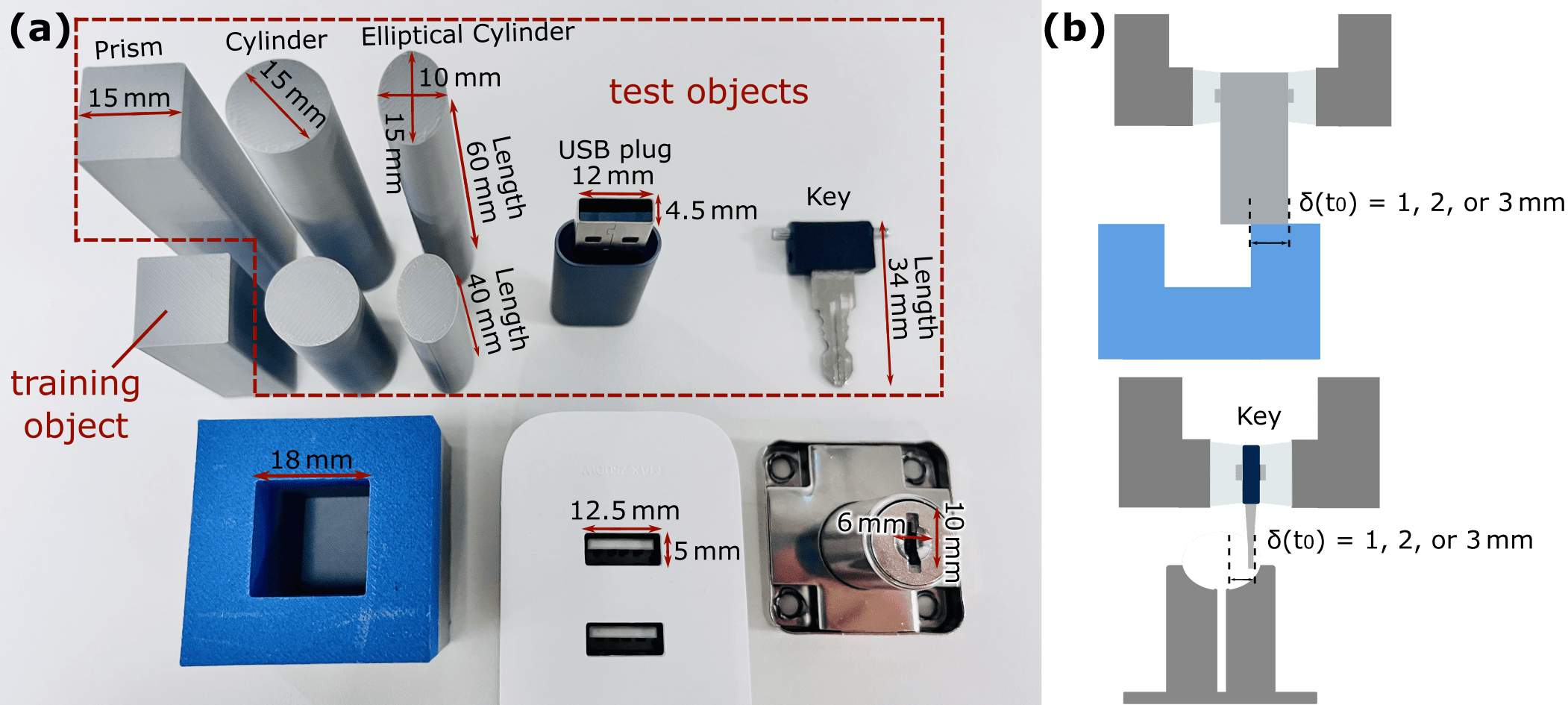}
    \caption{The experiment involves inserting objects into a rectangular slot that is 3\,mm wider than the objects. A 40\,mm prism was used as the training object, while the remaining ones are testing objects. The tasks include inserting 40\,mm and 60\,mm long prism, round and elliptical cylinders into the rectangular slot, inserting a USB plug into a hub, and inserting a metal key into a keyhole.}
    \label{fig:experiment_configuration}
    \vspace{-0.5em}
\end{figure}

Fig.~\ref{fig:experiment_configuration} shows the training and testing objects used in the insertion experiments. Six 3D-printed plastic objects in the shapes of prism, cylinder, and elliptical cylinder are used, each with lengths of 40\,mm and 60\,mm. In addition, the testing is generalized to everyday objects, including a USB connector-hub pair and a key-lock pair.

\begin{figure}[tp] 
    \centering
    \includegraphics[width=90mm]{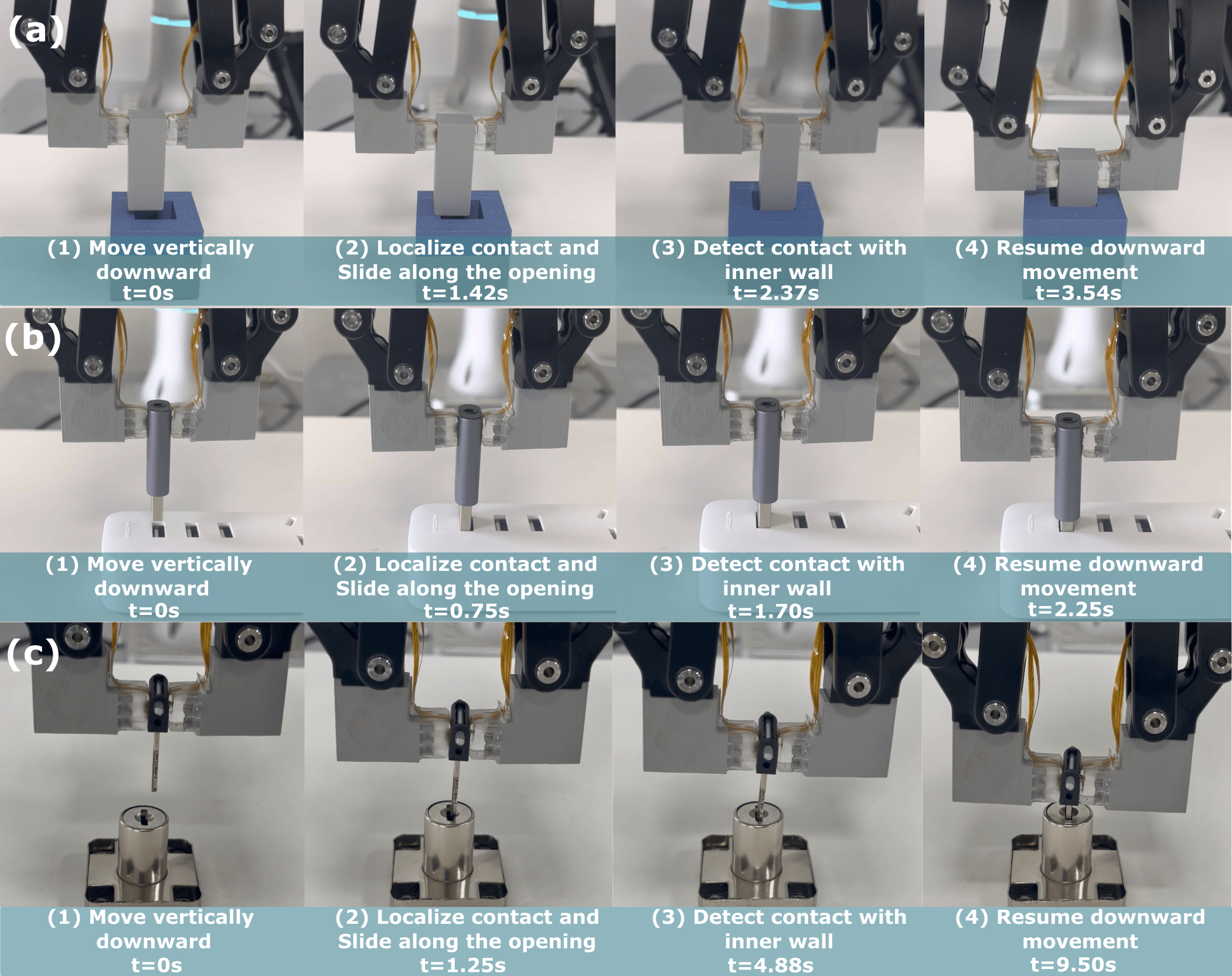}
    \caption{Physical tests with three types of insertion tasks without visual sensing. The initial position of the objects is randomized. Four insertion stages are shown.}
    \label{fig:task}
\end{figure}

The robot is trained to perform the insertion tasks mimicking human behaviors. The process consists of four stages (Fig.~\ref{fig:task}):

\textbf{Move vertically downward (Stage 1):} The gripper holding the object is positioned above the slot and moved vertically downward until contact is detected. This stage demonstrates the response speed of \finetact in contact detection.

\textbf{Localize contact and slide along the opening (Stage 2):} When contact occurs, the robot localizes the contact by analyzing subtle translational and torsional deformations of the fingertip induced by environmental disturbances, while simultaneously identifying the direction of the slot opening. As the deviation level increases, localizing the contact becomes more challenging due to the reduction of torsional deformations and translational differences between the two fingertips. The robot then slides the object along the opening using the \finetact sensors, and upon detecting a signal indicating departure from the edge, it incorporates a downward movement into its trajectory. Admittance control is applied to the robotic arm to ensure that contact forces remain within a safe threshold of 5\,N, using a proportional integral controller.

\textbf{Detect contact with inner wall (Stage 3):} \finetact can responsively track the dynamic 6-DoF poses, enabling it to promptly identify subtle contact with the inner wall and distinguish it from external contact outside the slot, as is common during stage 2. This stage demonstrates its ability to sense different contact modes.

\textbf{Resume downward movement (Stage 4):} This stage is relatively straightforward for the plastic prism-slot pair because the slot is 3\,mm wider than the plastic prism, allowing for greater angular misalignment during insertion. However, in real-world peg-in-hole insertion tasks, such as USB plug or key insertion, the clearance between the peg and the hole is much smaller, as shown in Fig.~\ref{fig:experiment_configuration}. Therefore, it is necessary for \finetact to sense the direction of obstruction and make appropriate adjustments.

To successfully accomplish this task, the robot must be able to perceive various contact modes and directions and take appropriate actions accordingly. We collect 40 demonstrations in total, with 10 demonstrations for each edge of the slot. Note that only the 40\,mm rectangular prism-slot pair is used for training, while the others are reserved for testing.

\subsubsection{Evaluation Metrics}
To ensure a fair comparison, the inserted object is initially offset from the target slot by 1 to 3 mm in each of the four lateral directions. For each deviation level, four lateral directions were tested and two rollouts were conducted per direction, resulting in a total of 24 rollouts per object. The task is considered successful if the object is adjusted towards the slot opening and successfully inserted into the slot.

\subsubsection{Qualitative Analysis}
\begin{table}[tp]
\centering
\caption{\textnormal{Success rates (out of 24) of insertion tasks using \finetact-only policy. Only the 40\,mm rectangular prism was used for training; the rest are unseen objects for generalization evaluation. Each object was tested under three deviation levels, represented by $\delta(t_0)$, which is the corrective movement distance required to complete an insertion task.}}
\label{tab:performance}
\scriptsize 
\setlength{\tabcolsep}{3pt} 
\renewcommand{\arraystretch}{1.1} 
\begin{tabularx}{\columnwidth}{lXXXX}
\toprule
Testing Objects & $\delta(t_0)=1$\,mm & $\delta(t_0)=2$\,mm & $\delta(t_0)=3$\,mm & Avg. Success Rate \\ 
\midrule
4\,cm rectangular prism   & 100\% & 87.5\% & 75.0\% & 87.5\% \\
4\,cm cylinder            & 75.0\% & 62.5\% & 62.5\% & 66.7\% \\
4\,cm elliptical cylinder & 87.5\% & 62.5\% & 50.0\% & 66.7\% \\
6\,cm rectangular prism   & 87.5\% & 75.0\% & 50.0\% & 70.8\% \\
6\,cm cylinder            & 75.0\% & 75.0\% & 50.0\% & 66.7\% \\
6\,cm elliptical cylinder & 75.0\% & 62.5\% & 25.0\% & 54.2\% \\
USB Plug                  & 75.0\% & 62.5\% & 50.0\% & 62.5\% \\
Metal Key                 & 87.5\% & 87.5\% & 87.5\% & 87.5\% \\
\bottomrule
\end{tabularx}
\end{table}
The quantitative results are presented in Table~\ref{tab:performance}, showing the success rates for inserting 3D printed objects, a USB plug and a metal key at different deviation levels. The 40\,mm rectangular prism was the only object used for training, achieving an average success rate of 88\% during testing. All other objects, used for generalization evaluation, still achieved an average success rate of nearly 70\% in insertion tasks. 
Our experimental results reveal several key insights. 
As the starting deviation distance $\delta(t_0)$ increases, the success rate decreases. This is because a larger deviation $\delta(t_0)$ causes the contact point to be closer to the rotational center, resulting in a shorter moment arm and thus a smaller torque. Consequently, the dynamic torsional deformations at the edge contact are reduced, leading to a weaker gyroscope signal and making direction discrimination more difficult. As shown in Fig.~\ref{fig:experiment_configuration}, the narrow edges of the elliptical cylinder and the USB plug are only 10\,mm and 4.5\,mm wide, respectively, which can lead to incorrect direction inference after contact. Given the minimal object dimensions, a 3 mm deviation already places the contact point near the grasp center of mass, resulting in negligible rotational and translational disturbances. Consequently, larger perturbations become feasible when grasping objects of increased size. Furthermore, the visuotactile policy experiments demonstrate that vision provides effective coarse localization (typically within 1–2 mm), after which tactile feedback enables precise fine-tuning through subtle corrective adjustments.
Collisions between objects of different shapes and the slot produce different contact modes, which in turn lead to varying dynamic translational and torsional deformations of the elastomeric tips, diminishing the success rate for inserting generalized objects. 
Experimental results with a USB plug and a metal key demonstrate that \finetact can be generalized to objects of various materials and geometries, as its sensing mechanism is based exclusively on directional deformation of the elastomeric tips.

\section{Limitations}
Although \finetact demonstrates its ability to leverage transient dynamic tactile information for completing contact-rich insertion tasks, several limitations remain.

\textbf{Limited sensing for spatial information:} 
Current design of \finetact prioritizes reducing sensing channels and maximizing the temporal resolution, but lacks the sensing capability of spatial patterns. 
One possible future direction is to increase the number of \gls{imu} integrated and to adapt super-resolution techniques. Existing research suggests that spatial information is partially encoded in the temporal structure of vibrotactile signals and thereby can be extracted via decoding methods \cite{shao2020compression}.

\textbf{Limited sensing for pseudo-static contact:} 
Since \finetact is centered on dynamic tactile sensing, it lacks the ability to measure constant and pseudo-static deformation. Thus, if it touches an object that slowly deforms the gripper tip, \finetact can barely detect any static pressure. As a result, integrating other types of sensing components, such as a magnetometer and corresponding magnetized elastomeric materials, is one direction to be explored in the future.

\textbf{Incomplete physical modeling:} The physical model describing the relationship between the 6D pose of the grasped object and the corresponding 6-axis IMU signals remains underexplored. Further research is needed to establish an explainable model for more robust pose estimation.

\textbf{Sensor size optimization:} 
The current design of elastomeric gripper tip is four times larger than the size of the \gls{imu} chip, suggesting that the size of the elastomeric tip can be further reduced with a more optimized mechanical design. A more compact gripper tip will allow the robot to manipulate objects of smaller sizes.

\textbf{Limited policy inference speed:} 
Although we have high sampling rate continuous tactile signal input, the iterative denoising process of the diffusion policy requires approximately 60\,ms per inference, which limits the robot's real-time feedback capabilities.
Further research is needed to explore algorithms that balance both inference speed and the ability to model multimodal action distributions.


\section{Conclusion} 
We present \finetact, a cost-effective tactile sensing and control framework designed to enable delicate robotic manipulation. 
We utilize a compact elastomeric gripper tip integrated with \gls{imu} to capture transient and subtle tactile signals arising from dynamic translational and torsional deformations of the tip when the grasped object has environmental contact. 
To infer extrinsic contact direction, tactile signals are encoded and fused with visual features, then fed into an action diffusion model. This establishes a sensing-to-control mapping between the 6-DoF deformation of the elastomeric gripper tips and the corresponding 6-DoF pose adjustments of the robot end-effector during insertion tasks
Our experiments demonstrate that \finetact, when used with a visuotactile policy, achieves a high average success rate of 79\% in grasping and insertion tasks. For downward insertion tasks specifically, it reaches 88\% success on the training object and nearly 70\% on average for unseen objects of varying shapes, sizes, and materials using the tactile-only policy.
The data efficiency, low cost, compact design, and robust performance of \finetact underscore its potential for advancing delicate robotic manipulation in contact-rich environments.

\bibliographystyle{plain}  
\bibliography{ref}         

\end{document}